\documentclass[nohyperref]{article}
\usepackage[utf8]{inputenc}
\usepackage{arxiv}

\usepackage{microtype}    % Slightly tweak font spacing for aesthetics

\usepackage{fullpage}
\usepackage{natbib}
\usepackage{algorithm,algorithmic}
\usepackage{amsmath,amssymb,mathtools}
\usepackage{bm}
\usepackage{amsthm}
\usepackage{graphicx,subfig,color,wrapfig,epsfig,epstopdf}
\usepackage{enumitem}
\usepackage{booktabs}
\usepackage{mathrsfs}
\newcommand\numberthis{\addtocounter{equation}{1}\tag{\theequation}}

\def\alg{MKQ-BERT}
\def\lb{\left\lfloor}
\def\rb{\right\rceil}
\def\x{\bm{x}}
\def\L{\mathcal{L}}
\def\O{\bm{O}}
\def\OA{\bm{OA}}
\def\softmax{\text{Softmax}}

\begin{document}

\title{{\alg}: Quantized BERT with 4bits Weights and Activations}
\author{{Hanlin Tang}\thanks{Corresponding to Hanlin Tang $<$ranchotang@tencent.com$>$} \\
	\texttt{ranchotang@tencent.com} \\
	Tencent AIPD\\
	%% examples of more authors
	\And
	{XiPeng Zhang} \\
	Tencent AIPD\\
	\texttt{xipengzhang@tencent.com} \\
    \And
	{Kai Liu} \\
	Tencent AIPD\\
	\texttt{raccoonliu@tencent.com} \\
	\And
	{Jianchen Zhu} \\
	Tencent AIPD\\
	\texttt{dickzhu@tencent.com} \\
	\And
	{Zhanhui Kang} \\
	Tencent AIPD\\
	\texttt{kegokang@tencent.com} \\
}

\maketitle

\begin{abstract}
Recently, pre-trained Transformer based language models, such as BERT, have shown great superiority over the traditional methods in many Natural Language Processing (NLP) tasks. However, the computational cost for deploying these models is prohibitive on resource-restricted devices. One method to alleviate this computation overhead is to quantize the original model into fewer bits representation, and previous work has proved that we can at most quantize both weights and activations of BERT into $8$-bits, without degrading its performance. In this work, we propose {\alg}, which further improves the compression level and uses $4$-bits for quantization. In {\alg}, we propose a novel way for computing the gradient of the quantization scale, combined with an advanced distillation strategy. On the one hand, we prove that {\alg} outperforms the existing BERT quantization methods for achieving a higher accuracy under the same compression level. On the other hand, we are the first work that successfully deploys the $4$-bits BERT and achieves an end-to-end speedup for inference. Our results suggest that we could achieve $\bm{5.3}$x of bits reduction without degrading the model accuracy, and  the inference speed of one int$4$ layer is $\bm{15}$x faster than a float$32$ layer in Transformer based model.
\end{abstract}

\section{Introduction}
Recent Transformer based language models (e.g. BERT~\citep{bert}) have achieved remarkable performance on many natural language processing tasks. While the performance of these large  models has significantly increased, the number of parameters of these models also grows dramatically. Hence the models are usually computation expensive and memory intensive.  Therefore how to achieve efficient, real-time models with optimal accuracy has become more and more important. Many approaches has been proposed to alleviate this problem. For example, automated searching for efficient model architectures using technique such as NAS~\citep{zoph2016neural}, parameter pruning that removes redundant weights in the model~\citep{lecun1990optimal,hassibi1993second}, knowledge distillation that transfer the knowledge from large models to smaller models in order to boost its performance~\citep{hinton2015distilling}. 

Beyond those methods, one important direction is model quantization, and it has already shown great success for achieving an end-to-end inference speedup.  Previous work has already proved that  Transformer based models can be quantized into $8$-bits representation without degrading its performance~\citep{q8bert}, and since the model architecture does not change, the only thing left is to design hardware that can support the operation for quantized int$8$ numbers. Although it seems to be a very promising approach, previous methods are limited to the case where we can at most quantize the numbers into $8$-bits without performance degradation, and when being extended to a fewer bits case, such as $4$-bits, the performance drops severely. 

Therefore in this paper, we focus on this important problem and proposed a novel method, so called {\alg}, that can improve the performance of quantized model for $4$-bits quantization, by using knowledge distillation and dynamic quantization. Inspired by previous work~\citep{kdlsq}, we find that knowledge distillation and Learned Step size Quantization (LSQ) can substantially improve the accuracy of the quantized model. Therefore we start from this point and designed new technique for these two components. More specifically, the contribution of our paper is summarized as follows:
\begin{enumerate}
    \item We designed a new algorithm for dynamically updating the quantization function. Instead of using a STE-based gradient for updating the quantization scale, which is widely adapted in previous work~\citep{kdlsq,esser2019learned}, we derived a new MSE-based gradient that can further reduce the quantization error and improve the accuracy of the quantized model.
    \item For knowledge distillation, we adapted an advanced distillation strategy that can not only achieve a higher testing results, but also can be easily applied to the case where the teacher model is deeper than the student.
    \item In order to evaluate the end-to-end improvement of {\alg}, we deigned efficient kernels for implementing the quantized $4$-bits matrix multiplication. To the best of our knowledge, this is the first work that successfully deploys an int$4$ quantized model.
\end{enumerate}
In order to evaluate the performance of {\alg}, We conduct  extensive  experiments  on  downstream NLP tasks. Experimental results demonstrate that {\alg} outperforms  state-of-the-art  baseline, and has achieved almost the same performance with the full precision model when $\bm{50}\%$ layers being quantized into $4$-bits with the rest layers being quantized into $8$-bits. Moreover, our experimental results indicates that by using $4$-bits quantization, we could achieve $\bm{15}$x of end-to-end inference speedup for one Transformer layer.

\section{Related Work}\label{sec:related}
\paragraph{Efficient NN inference:} Due to the high demand for decreasing the computation cost of large NNs, there has been a large body of literature that focused on designing method to improve the inference speed of NN models, without degrading its performance. One line of those tryouts is to design efficient model architecture, either manually~\citep{howard2019searching,tan2019efficientnet}, or automatically using Automated machine learning (AutoML) and Neural Architecture Search (NAS). The automated searching methods is more powerful in finding the right architecture given certain constrains of computation cost~\citep{wu2019fbnet,wang2020attentivenas,tan2019mnasnet,elsken2019neural}. 

Meanwhile, since most of the NNs are over-parameterized, therefore it usually contains a lots of unimportant parameters that can be removed, and this method is so called model pruning. In pruning, neurons that is less important is safely removed, without sacrificing the model accuracy. The pruning methods can be roughly categorized into unstructured pruning~\citep{lee2018snip, xiao2019autoprune, park2020lookahead} and structured pruning~\citep{yu2018nisp, lin2018accelerating, huang2018data, zhao2019variational,yu2021hessian}. 

Another important direction for reducing the model size is knowledge distillation~\citep{polino2018model, ahn2019variational, yin2020dreaming}. The main idea of knowledge is to use a well trained large model as teacher to train a more compact model. Instead of using the label from the training data solely, it would also utilize the output from the teacher since it usually contain more information than the training data. Notice that knowledge does not change the model structure, therefore it can be easily combined with other techniques, such as pruning and model quantization, to further improve the performance of the compact models~\citep{mccarley2019pruning,stock2019and,polino2018model}.

\paragraph{Model quantization:} In model quantization, the parameters and activations of the model are represented using fewer bits~\citep{lin2015neural,hubara2016binarized,rastegari2016xnor, gysel2016hardware, gysel2018ristretto, tailor2020degree, ni2020wrapnet}. However, directly quantizing the model will greatly decrease the accuracy of the models, therefore there has been many attempts to improve the performance of the quantized model. One important technique is Quantization Aware Training (QAT)~\citep{jacob2018quantization}, where it simulates the quantization procedure in training to further improve the accuracy of the quantized model. 

For Transformer based models, the boundary of the compression level has been continuously advanced. For example, 8-bit quantization training is has been successfully applied in FullyQT~\citep{prato2019fully} and Q8BERT \cite{zafrir2019q8bert}. (Fan et al., 2020) quantizes a different subset of weights in each training iteration to make models more robust to quantization. For even fewer bits quantization, Q-BERT \cite{shen2020q} achieves ultra-low bit quantization by using mixed-precision quantization bits. In Quant-Noise (QN) \cite{fan2020training}, authors proposed a novel method for attaining an unbiased estimation of the gradients by only quantizing a subset of weights in each iteration. For tenary case, TernaryBERT \citep{zhang2020ternarybert} divides the original model into separate parts  through 2-bit ternarization training. The work that is most similar to our method is KDLSQ-BERT~\citep{kdlsq}, where authors combined Learned Step size Quantization (LSQ) with knowledge distillation. However, they only consider the case where the activations are quantized into $8$-bits with the weight being quantized into fewer bits (e.g., $4$-bits or $2$-bits), this type of design is not hardware friendly therefore can not be implemented efficiently to get an end-to-end inference throughput improvement.

\section{Preliminary}\label{sec:background}
In this section, we present some preliminary knowledge about model quantization, Transformer network and knowledge distillation.
\subsection{Quantization}\label{sec:calibration}
One solution to reduce the computation cost is to replace the original float32 numbers  into fewer-bits (i.e. int8) representation during computation, and this is so called model quantization. Generally speaking, we can quantize a float number $x$ into $k$-bits representation according to
\begin{align*}
Q[x] = s\times\text{round}\left(\text{clamp}\left(\frac{x}{s},l_{\min},l_{\max}\right)\right).\numberthis\label{eq:def_q}
\end{align*}
Here $s$ is the quantization scale, $l_{\min}$ and $l_{\max}$ are the lower and upper bound for the clamping function, the rounding function (denoted as $\lb\cdot\rb$) will round the number into its nearest integer. Notice that for $k$-bits quantization, usually we would set $l_{\min} = \left(-2^{k-1}+1\right)$ and $l_{\max} = 2^{k-1}$. For the case where $x$ is a tensor $\x$, previous work either uses a common $s$ for all number (per-tensor scale) or only the elements within the same row share the same scale (per-row scale).

From \eqref{eq:def_q} we can see that, if $s$ is smaller, the rounding interval will shrink, which means the quantization error $|Q[x] - x|$ also decays for $x\in[l_{\min},l_{\max}]$. However, if $s$ is too small, the number of elements that exceed the clamping bound would also increase, which makes the overall quantization error $\|Q[\x] - \x\|$ increase. Therefore finding a good balance for $s$ is every essential for reducing the side effect of quantization. 

In \citet{q8bert}, authors uses the maximum of the absolute value for each weight tensor as the quantization scale; for activation, since it will change with different input, therefore they first sample a few portion of the training data, and then get the empirical statistical distribution of the activation's absolute value for those samples. The use the top $0.01\%$ largest value as the initial scale. This procedure for setting the initial value of the quantization scale  is so called calibration. After calibration, The quantized mod could get further improved by QAT~\citep{jacob2018quantization}. Previous work has already proved that by combining calibration with QAT, using $8$-bits quantization, we can quantize BERT without performance degradation.

However, when further decreasing the representation bits to $k=4$, the performance degrades severely, and for $4$-bits quantized model, they requires more QAT steps. The problem is that when training steps increases, the deviation between the trained model and  the initial model also increase, therefore using original quantization scale is not optimal in this case. In order to capture the trend of the model weight updates, previous work \citep{kdlsq} proposed a learned step size quantization (LSQ) method to dynamically updating the quantization scale, which will be detailed discussed in Section~\ref{sec:method}.

\subsection{Backbone Network: Transformer}
\label{sec:transformer}

In transformer, denotes the $l$-th layer as $\mathrm{Transformer_l(\cdot)}$, and $\bm{h}^{0}\in\mathbb{R}^{d_h}$ as the input tensor $\x$, then the  stacked Transformer blocks compute the encoding vectors according to:
\begin{equation}
\bm{h}^l = \mathrm{Transformer}_{l}(\bm{h}^{l-1}),~l \in [1, L]
\end{equation}
where $L$ is the number of Transformer layers. Each layer  consists of a self-attention layer and a fully connected feed-forward network. 

\paragraph{Multi-Head Attention (MHA)} The self-attention layer consists of multiple self-attention heads. Assuming that there are $A_h$ heads, each attention head has three components: Query matrix $\bm{Q}_{l,a}$, Key matrix $\bm{K}_{l,a}$ and Value matrix $\bm{V}_{l,a}$. The output of each attention head (denoted as $\OA_{l,a}$) is computed via:
\begin{gather}
\bm{q}_{l,a} = \bm{h}^{l-1} \bm{Q}_{l,a},~\bm{k}_{l,a} = \bm{h}^{l-1} \bm{K}_{l,a},~\bm{v}_{l,a} = \bm{h}^{l-1} \bm{V}_{l,a}, \\
\bm{A}_{l,a} = \softmax\left(\frac{\bm{q}_{l,a} \bm{v}_{l,a}^{\intercal}}{ \sqrt{d_k}}\right),  \\
\bm{OA}_{l,a} = \bm{A}_{l,a}\bm{v}_{l,a},
\end{gather}
where $\bm{A}_{l,a} \in \mathbb{R}^{|x| \times |x|}$ is a matrix that indicates the attention distributions and $d_k = \frac{d_h}{A_h}$. Afterwards, all $\OA_{l,a}$ are concatenated together and fed into one fully connected (fc) layer, which can be expressed as
\begin{align*}
\OA_l = [\OA_{l,1},\cdots,\O_{l,A_h}]\bm{W}^A + \bm{b}^A,
\end{align*}
where $\bm{W}^A$ and $\bm{b}^A$ are the weight and bias of the fc layer and $\OA_l$ is the output of MHA.
\paragraph{Feed-Forward Network (FFN)} The feed-forward network consists of two fully connected network interconnected by one activation function. The output of the self-attention layer is treated as the input to the FFN, the output of FFN is computed via
\begin{align*}
\bm{FFN}(\x) = \text{GELU}(\x\bm{W}^1 + \bm{b}^1)\bm{W}^2 + \bm{b}^2, 
\end{align*}
where $\bm{W}^1$ and $\bm{W}^2$ are the weight of the fc layer, $\bm{b}^1$ and $\bm{b}^2$ are the bias respectively.

\subsection{Knowledge Distillation}
The basic idea of knowledge distillation is to transfer the knowledge from a powerful model (teacher model) into a weaker model (student model, usually admits a smaller size or few bits than the teacher model). This is done by letting the student model's features to mimic the features from the teacher model, which can be achieved by minimizing the difference between these two sets of features:
\begin{align*}
\mathcal{L}_{\text{KD}} = {\sum_{e\in \mathcal{D}}{L(f^{S}(e), f^{T}(e))}},\numberthis \label{eq:kd_output}
\end{align*}
where $\mathcal{D}$ denotes the training data, $f^{S}(\cdot)$ and $f^{T}(\cdot)$ indicate the features from the student and teacher models, $L(\cdot)$ is the loss function that measures the difference of the features. Previous work treat the uncompressed model as the teacher model and the quantized model as student, and they used two sets of features for distillation:
\begin{itemize}
\item \textbf{Output distillation:} The output of the network is treated as the distillation feature, and the loss can be either mean square error (MSE) or KL-divergence. This loss can be expressed as $\L_{output} = L(\bm{O}^S,\bm{O}^T)$, where $\bm{O}^S$ and $\bm{O}^T$ are the outputs from the student model and teacher model;
\item \textbf{Attention distillation}: The output of the self-attention network and the feed-forward network are used for distillation. More specifically, two sets of outputs $\{ \bm{A}_{l,a} \}$ and $\{\O_{l,a}^A\}$ for each attention head and attention layers, which can be written as
\begin{align*}
&\L_{attention}\\
=& \sum_{a}\sum_l \left(L(\bm{A}_{l,a}^S,\bm{A}_{l,a}^T) +  L(\OA_{l,a}^S,\OA_{l,a}^T)\right).\numberthis \label{eq:kd_attention}
\end{align*}
\end{itemize}
In ~\citep{kdlsq}, authors add \eqref{eq:kd_output} and \eqref{eq:kd_attention} together with the original training loss $\L_{train}$ into $\L_{final} = \L_{output} + \L_{attention} + \L_{train}$, and use this $\L_{final}$ as the final training loss. However, those method 
can only use teacher models with the same model configuration, which means we cannot use a larger network to further improve the performance of the quantized model.

\section{Methodology}\label{sec:method}
The pipeline of {\alg} can be divided into two steps: calibration step and QAT step. The calibration step is used to get a good initial value for the quantization scale and we follow the same strategy as described in Section~\ref{sec:calibration}. Afterwards, we start QAT  to finetune both the model parameter and the quantization scale. For QAT,  we further improve the previous QAT methods using the following two strategies:
\begin{itemize}
    \item \textbf{New gradient for quantization scale}: In order to find a better scale factor, we designed a new strategy for the computing the gradient of the quantization scale, and we use this gradient to dynamically updating the quantization scale.
    \item \textbf{Advanced distillation strategy}: In {\alg}, we use an advanced distillation strategy that can not only achieve a higher testing results, but also can be easily applied to the case where the teacher model is deeper than the student.
\end{itemize}

\subsection{New algorithm for dynamic quantization}
Instead of adapting the same strategy from previous work~\citep{kdlsq,esser2019learned} for updating the quantization function, we propose a new method for computing the gradient of the quantization scale.

\subsubsection{STE-based gradient for quantization scale}
Specifically, previous work attains this gradient using STE (Straight Through Estimation) according to
\begin{align*}
\frac{\partial Q[x]}{\partial s} =& \frac{\partial\left(\lb\frac{x}{s}\rb s \right)}{\partial s}\\
= &s\frac{\partial\left(\lb\frac{x}{s}\rb  \right)}{\partial s} + \lb\frac{x}{s}\rb\frac{\partial\left(s \right)}{\partial s}\\
= & s\frac{\partial\left(\frac{x}{s}  \right)}{\partial s} + \lb\frac{x}{s}\rb\quad\left(\frac{\partial \lfloor\cdot\rceil}{\partial s} = 1 \right)\\
=& -\frac{x}{s} + \lb\frac{x}{s}\rb.
\end{align*}
For the case where $x$ becomes a tensor $\x$, the gradient would be
\begin{align*}
\frac{\partial Q[\x]}{\partial s} = \sum_{i}\left(-\frac{x_i}{s} + \lb\frac{x_i}{s}\rb \right),
\end{align*}
where $x_i$ is the $i$-th component of $\x$. In the backward procedure, previous work define $\frac{\partial f}{\partial s} = \frac{\partial Q[\x]}{\partial s}$ and uses this gradient for updating the quantization scale.

However, our analysis indicates that this method is not optimal. Consider the case below: if we have a tensor $\x = (0.2, 0.9)$ and the quantization scale $s=1$, which means the quantized version of $\x$ equals to 
\begin{align*}
  Q[\x] = 1\times(0,1) = (0,1).  
\end{align*}
 For this tensor, if we decrease $s$ into $0.9$, then $Q[\x] = (0,0.9)$, it will leads to a better quantization choice than the result from $s=1$. However, the gradient attained from STE admits
\begin{align*}
\frac{\partial Q[\x]}{\partial s} = -0.2 + 0.1 = -0.1.
\end{align*}
Since the gradient is negative, in the next step we are going to increase $s$. But as shown above, what we need is to decrease $s$, this is not consistent with the gradient from STE, and this motivates us to design a new method for approximating the gradient.

\subsubsection{MSE-based gradient for quantization scale}
Here we propose a MSE-based gradient for the quantization scale. In order to minimize the quantization error w.r.t. the whole tensor $\x$. Let's say the original scale $s$ gets an infinitely small variation $\Delta s$, which means 
\begin{align*}
\lb \frac{x}{s + \Delta s}\rb = \lb \frac{x}{s}\rb, \quad\text{if } \frac{x}{s}- \lb \frac{x}{s + \Delta s}\rb\neq 0.5.
\end{align*}
Therefore we get
\begin{align*}
&Q_{s+\Delta s}[x]\\
= &(s+\Delta s)\lb \frac{x}{s + \Delta s}\rb\\
=& (s+\Delta s)\lb \frac{x}{s}\rb,
\end{align*}
this leads to
\begin{align*}
\frac{\partial Q[x]}{\partial s} = \frac{Q_{s+\Delta s}[x] - Q_{s}[x]}{\Delta s} = \lb \frac{x}{s}\rb.
\end{align*}
Notice that for MSE-based gradient, the major goal is to minimize the quantization error $\|Q[\x] - \x\|^2$, not to estimate the gradient of the quantization scale w.r.t. the quantization function ($\frac{\partial Q[x]}{\partial s}$), therefore we redefine the gradient of the quantization scale as
\begin{align*}
\text{Gradient}(s):=&\frac{\partial(Q[x] - x)^2}{\partial s}\\
= &2(Q[x] - x)\frac{\partial Q[x]}{\partial s}\\
=& 2 (Q[x] - x)\lb \frac{x}{s}\rb,
\end{align*}
which gives us 
\begin{align*}
\frac{\partial(Q[\x] - \x)^2}{\partial s}= 2\sum_{i}\left((Q[x_i] - x_i)\lb \frac{x_i}{s}\rb\right)
\end{align*} and we define 
\begin{align*}
\frac{\partial f}{\partial s} := \text{Gradient}(s) = \frac{\partial(Q[\x] - \x)^2}{\partial s}.
\end{align*}

Back to case before, in this case, the MSE-based gradient equals to 
\begin{align*}
\frac{\partial(Q[x] - x)^2}{\partial s} = 2\left(-0.2 \times 0 + 0.1*1 \right) = 0.2,
\end{align*}
which is a positive number, so we will decrease $s$ in the next step, which explains why MSE-based gradient is better than STE-based gradient.

\subsection{MINI distillation with different scale factor}
In \citet{mini}, authors found that by only using the output of MHA and FFN from the last layer can even outperform the performance of multi-layer distillation. More importantly, this distillation method is applicable for the case where the teacher model is deeper than the student without manually specifying the layer correspondence, which means it can be easily extend to the case where we want to use a larger model to teach the quantized model. For the output of MHA, it defines
\begin{align*}
\L_{attention} = \sum_{a}KL(\OA_{a}^S\|\OA_{a}^T),\numberthis\label{eq:kd_attention_mini}
\end{align*} where $KL(\bm p_1|| \bm p_2)$ indicates the KL-divergence of two distributions $\bm p_1$ and $\bm p_2$. Unlike previous work which uses the output of FFn for distillation, it uses value vectors. More specifically, it first fed $\bm{v}_{l,a}$ into a softmax layer, then use KL divergence to measure the distance of  softmax layer's results   between  the student and teacher model, which can be written as
\begin{align*}
\hat{\bm{v}}_{l,a}^S =& \softmax\left(\frac{v_{l,a}^S\left(\bm{v}_{l,a}^S \right)^\top}{\sqrt{d_k}}\right)\\
\hat{\bm{v}}_{l,a}^T =& \softmax\left(\frac{v_{l,a}^T\left(\bm{v}_{l,a}^T \right)^\top}{\sqrt{d_k}}\right)\\
\L_{value}=&\sum_{a}KL(\hat{\bm{v}}_{l,a}^S\|\hat{\bm{v}}_{l,a}^T),\numberthis\label{eq:kd_value_mini}
\end{align*}
where $\hat{\bm{v}}_{l,a}^S$ and $\hat{\bm{v}}_{l,a}^T$ are the value vector from the student and teacher model for each attention head.

In {\alg}, we adapt the this distillation strategy and combine \eqref{eq:kd_output}, \eqref{eq:kd_attention_mini}, \eqref{eq:kd_value_mini} together with the original training loss with different scale factor as the final training loss:
\begin{align*}
\L_{final} = \L_{train} + \alpha \L_{output} + \beta\left(\L_{attention} + \L_{value} \right).\numberthis\label{eq:kd_alpha_beta}
\end{align*}
Empirically, we find that by setting $\alpha$ larger than $\beta$ (e.g. $\alpha=10$ and $\beta=0.5$) is beneficial for the student to achieve a better performance. 

\begin{table*}[t]
  \caption{GLUE development set results. TinyBERT4 (original) are the fintuned results using checkpoint from \citet{tinybert}. TinyBERT4$_{4}$ are the results using {\alg} with the $4$th layer quantized into $4$-bits. TinyBERT4$_{3,4}$ are the results using {\alg} with the $3$rd and $4$th layer quantized into $4$-bits. TinyBERT4$_{2,3,4}$ are the results using {\alg} with the $2$nd, $3$rd and $4$th layer quantized into $4$-bits. TinyBERT4$_{1,2,3,4}$ are the results using {\alg} with all layers except the embedding layer being quantized into $4$-bits. All layers (except embedding) use $8$-bits quantization as default setting if not being quantized into $4$-bits. All models ends with  (KDLSQ) means using the method from KDLSQ-BERT~\citep{kdlsq}.}\label{table:glue}
  \centering
  \begin{tabular}{lccccccc}
  \hline  %添加表格头部粗线
  \textbf{Model}& RTE& MRPC& CoLA & SST-2& QNLI& QQP \\
  \hline  %添加表格中横线
  TinyBERT4 (original) & 67.5 & 85.3 & 69.4  & 90.4 & 85.4& 87.1\\
  \hline
  TinyBERT4$_{4}$ & 68.5 & 85.0 & 70.2  & 90.4 & 85.3& 86.8\\
  TinyBERT4$_{4} (KDLSQ)$ & 67.3 & 84.0 & 69.7  & 89.6 & 84.9& 86.3\\
  \hline
  TinyBERT4$_{3,4}$ & 67.5 & 84.5 & 70.6  & 90.4 & 85.0& 86.7\\
  TinyBERT4$_{3,4} (KDLSQ)$ & 66.4 & 82.1 & 69.6  & 87.1 & 84.0& 86.1\\
  \hline
  TinyBERT4$_{2,3,4}$ & 67.1 & 81.3 & 70.1  & 88.3 & 74.4& 71.3\\
  TinyBERT4$_{2,3,4} (KDLSQ)$ & 56.3 & 71.3 & 69.4  & 76.6 & 69.8& 66.4\\
  \hline
  TinyBERT4$_{1,2,3,4}$ & 58.4 & 69.1 & 69.1  & 78.6 & 56.3& 66.3\\
  TinyBERT4$_{1,2,3,4} (KDLSQ)$ & 66.4 & 64.8 & 69.3  & 53.5 & 59.1& 61.2\\
  \hline %添加表格底部粗线
  \end{tabular}
\end{table*}
\section{Experiments}
In order to evalute the performance of {\alg}, we conduct model quantization experiments using different model quantization strategies, and evaluate the quantized models on the GLUE benchmark. Notice that KDLSQ~\citep{kdlsq} is the only work that combines learned step size quantization with knowledge distillation for Transformer based models, and it has already outperformed other methods on GLUE for $8$-bits quantization and $4$-bits weight quantization (the activations are still quantized into $8$-bits, which is not the case in our implementation). Therefore we only choose KDLSQ for comparison. All layernorm and activation functions are computed using float32 numbers in order to improve the model's performance.

\subsection{GLUE} 
The General Language Understanding Evaluation (GLUE) benchmark~\cite{wang2018glue} consists of several sentence-level classification tasks, including Corpus of Linguistic Acceptability (CoLA)~\cite{cola2018}, Stanford Sentiment Treebank (SST)~\cite{sst2013}, Microsoft Research Paraphrase Corpus
(MRPC)~\cite{mrpc2005},  Quora Question Pairs (QQP)~\cite{chen2018quora}, Question Natural Language Inference (QNLI), Recognizing Textual Entailment (RTE)~\cite{rte1}.

\subsection{Quantization Setup}
We use a TinyBERT4~\citep{tinybert} model (the number of layers $M=4$, the hidden size $d_h=312$, the intermediate size $d_i=1200$ and the head number $A_h=12$) that has a total of 14.5M parameters. We use the checkpoint released from TinyBERT as encoder, and finetune TinyBERT for each downstream task. The batch size is $32$ and learning rate is chosen from $\{1e-5,\, 3e-5,\, 5e-5\}$. After finetuning, we use the checkpoint that has the highest development accuracy for quantization. 

For calibration, we run the forward pass for $200$ steps with batch size 32. After the quantization being initialized, we would start the QAT. We use Adam as the optimizer. The learning rate for weight is chosen from $\{5e-6,\,1e-5,\,5e-5\}$, the learning rate for ativations' quantization scale is chosen from $\{0.05,\, 0.01\}$, the learning rate for weights' quantization is chosen from $\{0.005,\,0.001\}$. All learning rates follow the same scheduler that grows linearly for $10\%$ of the training steps and decays to $0$ till the end. We set $\alpha = 10$ and $\beta=1$ in \eqref{eq:kd_alpha_beta} for distillation. The max sequence length is 128 for each task. Empirically, we find that the higher levels are more robust to quantization therefore we start from the last layer for quantization. We run QAT for $30$ epochs for each task and report the best result over all hyper parameters.

\subsection{Main Results}

The development results for each algorithm are listed in Table~\ref{table:glue}. We shall see that by using {\alg}, we could achieve almost the same performance with the uncompressed model even when $50\%$ of layers being quantized into $4$-bits. When quantizing $75\%$ layers into $4$-bits, the model being trained using {\alg} still achieves a comparable performance with the original model for most tasks. However, when quantizing all layers into $4$-bits, the performance drops severely, this leads to a challenging problem for future studies. All results except four int$4$ layers case from {\alg} substantially outperforms the results from KDLSQ.

\subsection{End-to-end Inference Speedup}
As mentioned, another major contribution of our work is that we successfully implemented int$4$ matrix multiplication with efficient CUDA kernels.  Since the overall inference time depends on the number of int$4$ layers in the model, so here we report the averaged inference time for one layer, using int$4$ quantization, int$8$ quantization and float$32$ number for comparison. From Table~\ref{table:e2e} we shall see that when quantizing the layers into $4$-bits, it is $1.25$x faster than the int$8$ layer, and is even $15$x faster than the float32 layer. 
\begin{table}[]
  \caption{End-to-end inference time for running one layer in BERT-base model with different batch size and different numbers of valid tokens. We run our experiments on NVIDIA T4 GPUs, and the inference time is averaged over 100 rounds of experiments.}\label{table:e2e}
  \centering
  \begin{tabular}{lccccccc}
  \hline  %添加表格头部粗线
  BS & valid tokens & float32 (us) & int$8$ (us)& int$4$ (us)  \\
  \hline  %添加表格中横线
  16 & 440& 1380 & 213.1   & 160.5 \\
  16  & 537& 1845 & 245.7   & 179.3\\
  16 & 681& 2690 & 260.9   & 196.5 \\
  \hline
  64 & 1691& 6398 & 567.4   & 428.8 \\
  64 & 2011& 7185 & 628.4   & 490.1\\
  64 & 2298& 7897 & 669.9   & 533.4 \\
  \hline %添加表格底部粗线
  \end{tabular}
\end{table}

\begin{table*}[]
  \caption{GLUE development set results for ablation studies. TinyBERT4$_{3,4}$ are the results using {\alg} with the $3$rd and $4$th layer being quantized into $4$-bits. TinyBERT4$_{3,4}$ (w/o MINI KD) are the results using {\alg} without $\L_{attention}$ and $\L_{value}$ in \eqref{eq:kd_alpha_beta}. TinyBERT4$_{3,4}$ (w/o output KD) are the results using {\alg} without $\L_{output}$ in \eqref{eq:kd_alpha_beta}. TinyBERT4$_{3,4}$ (w/o LSQ) are the results using {\alg} with the quantization scale held unchanged along the training.}\label{table:ablation}
  \centering
  \begin{tabular}{lccccccc}
  \hline  %添加表格头部粗线
  \textbf{Model}& RTE& MRPC& CoLA & SST-2& QNLI& QQP\\
  \hline  %添加表格中横线
  TinyBERT4$_{3,4}$ & 67.5 & 84.5 & 70.6  & 90.4 & 85.0& 86.7\\
  TinyBERT4$_{3,4}$ (w/o MINI KD) & 66.0& 83.5 & 69.0   & 89.7 & 84.1& 86.1\\
  TinyBERT4$_{3,4}$ (w/o output KD) & 66.7& 83.8 & 70.0   & 90.0 & 84.4& 86.4\\
  TinyBERT4$_{3,4}$ (w/o LSQ) & 66.0& 84.0 & 69.7   & 90.0 & 83.6& 85.8\\
  \hline %添加表格底部粗线
  \end{tabular}
\end{table*}

\subsection{Ablation Studies}
In order to get a clear understanding for each component in {\alg}, we perform ablation studies to evaluate the performance of the quantized model without each components. There are three major components in {\alg}:
\begin{enumerate}
    \item MINI knowledge distillation;
    \item Output layer knowledge distillation;
    \item Learned step size quantization.
\end{enumerate}
Since we can at most quantize $50\%$ of the layers into $4$-bits, therefore we compress the $3$rd and $4$th layers of TinyBERT4 into $4$-bits with the rest layers being quantized into 8bits. The training setup still remains the same and we report the best result for each algorithm. As we can see that, for QNLI, which has more data samples than the other tasks, LSQ becomes more important because the quantized model would deviate from the initial model further since it runs for more training steps. The other trend is that MINI distillation is more important than the output distillation, we think this might be because it has already contains enough information of the encoder network. All of those results suggest that {\alg} is the best option.

\section{Conclusion}
In this work, we provide a novel quantization pipeline, named {\alg}, that  further improves the model's performance when quantizing the model weights and activations into $4$-bits. We propose a more robust way of dynamically updating the quantization function, and is proved to be very essential for reducing the side effect of the quantzation error. Beyond this, we apply an advanced distillation technique that outperforms the previous layer-to-layer attention distillation in QAT. Empirical studies has proved that {\alg} can substantially outperforms  previous QAT methods. Last but not least, we implement this int$4$ Transformer layer efficiently and  achieved an end-to-end inference throughput improvement. In the future we believe it would be very interesting and challenging to further extend the compression level without performance degradation.

% \newpage
\bibliography{iclr2021_conference}
\bibliographystyle{iclr2021_conference}

\end{document}